\documentclass[journal,12pt,onecolumn,draftclsnofoot,]{IEEEtran}
\IEEEoverridecommandlockouts
\usepackage{cite}
\usepackage{amsmath,amssymb,amsfonts}
\usepackage{graphicx,float}
\usepackage{textcomp}
\usepackage{xcolor}
\usepackage{csquotes}
\usepackage{subcaption}
\usepackage{multicol}
\usepackage{lipsum}
\usepackage[bookmarks=false]{hyperref}
\usepackage{tabularx,multirow,booktabs,blindtext}
\def\BibTeX{{\rm B\kern-.05em{\sc i\kern-.025em b}\kern-.08em
    T\kern-.1667em\lower.7ex\hbox{E}\kern-.125emX}}

\usepackage{authblk}
\usepackage{algorithm}
\usepackage{algpseudocode}
\algblock{Input}{EndInput}
\algnotext{EndInput}
\algblock{Output}{EndOutput}
\algnotext{EndOutput}
\newcommand{\Desc}[2]{\State \makebox[3em][l]{#1}{#2}}

\newcommand*{\affaddr}[1]{#1} 
\newcommand*{\affmark}[1][*]{\textsuperscript{#1}}
\newcommand*{\email}[1]{\textit{#1}}

\begin{document}

\title{A Comparative Study of Detecting Anomalies in Time Series Data Using LSTM and TCN Models}


\author{%
Saroj Gopali\affmark[1], Faranak Abri\affmark[1], Sima Siami-Namini\affmark[2], and Akbar Siami Namin\affmark[1]\\
\affaddr{\affmark[1]Department of Computer Science,}
\affaddr{\affmark[2]School of Planning and Public Policy} \\
\affaddr{\affmark[1]Texas Tech University,}
\affaddr{\affmark[2]Rutgers University} \\
\email{\{saroj.gopali $\vert$ faranak.abri $\vert$ akbar.namin\}@ttu.edu and sima.siaminamini@rutgers.edu}\\
}

\maketitle

\begin{abstract}
There exist several data-driven approaches that enable us model  time series data including traditional regression-based modeling approaches (i.e., ARIMA). Recently, deep learning techniques have been introduced and explored in the context of time series analysis and prediction. A major research question to ask is the performance of these many variations of deep learning techniques in predicting time series data. This paper compares two prominent deep learning modeling techniques. The Recurrent Neural Network (RNN)-based Long Short-Term Memory (LSTM) and the convolutional Neural Network (CNN)-based Temporal Convolutional Networks (TCN) are compared and their performance and training time are reported. According to our experimental results, both modeling techniques perform comparably having TCN-based models outperform LSTM slightly. Moreover, the CNN-based TCN model builds a stable model faster than the RNN-based LSTM models. 

\end{abstract}

\begin{IEEEkeywords}
Temporal convolutional network (TCN), Long Short-Term Memory (LSTM), Anomaly detection.
\end{IEEEkeywords}

\vspace*{-0.08in}
\section{Introduction}
\label{sec:intro}

With the everyday advances and innovations in connecting smart devices together and building a network of Internet of Things (IoT), it is expected that more home, manufacturing, and business applications embrace this emerging technology.
While the economic impact of IoT systems is soaring, there are some daunting issues and challenges that hurdles widespread use of this technology. One of the major problems is the security of IoT applications. According to a study \cite{IoT7}, in near feature more than 25\% of all attacks in enterprises will involve IoT systems making security as a top barrier to the successful deployment of IoT systems. 


A typical IoT system involves a myriad of various types of sensors with different quality and functions. These sensors collect and send data captured in surrounding environment. On the other hand, due to cost associated with physically and digitally securing these sensors, these small circuits are left without adequate security controls and thus they are prone to several cyber attacks including DoS attacks and tampering. 

There are several mathematically and data-driven models to detect anomalies in streaming data produced by IoT's sensors. Due to the temporal characteristics involved in such data, time series analysis and adapting state-of-the-art anomaly detection techniques in these types of data are quite prevalent in research and practice. For instance, machine learning techniques such as K Nearest Neighborhood (KNN), Feature Bagging (FB) and deep learning approaches such as Long Short-Term Memory (LSTM) and Graph-based learning have been adapted to anomaly detection in IoT time series data with reasonable performance and accuracy \cite{IoT19}. 

In this paper, we study the performance of two prominent deep learning approaches: The RNN-based Long Short-Term Memory (LSTM) and the CNN-based Temporal Convolutional Networks (TCN). TCN models are convolutional models capable of taking into account the temporal properties.
Unlike RNNs, convolutional-based models such as TCNs are believed to be less expensive yet produce comparative results. 
 The performance of the models we created was evaluated through a case study using water treatment testbed \cite{IoT17}. To make the comparison unbiased and meaningful, we built both models in a similar fashion by employing a similar number of hidden layers and complexity. Our results show that both TCN and LSTM produce comparable results in terms of predictions. However, TCN-based models outperforms slightly better than LSTM in terms of anomaly detection. Furthermore, TCN-based model outperforms slightly better in terms of training time. 
The key contributions of this paper are as follows:
\begin{itemize}
    \item[--] We compare the performance of multivariate RNN-based LSTM and CNN-based TCN models in the context of anomaly detection in time series.
    \item[--] We report that TCN-based models perform slightly better than TCN-based models in terms of prediction accuracy.
    \item[--] We also noticed that TCN-based models build stable models faster compared to the LSTM-based models. 
\end{itemize}

The rest of the paper is organized as follows: Section \ref{sec:relatedWork} reviews the related work. The technical background of TCN and LSTM are presented in Section \ref{sec:background}. Section \ref{sec:algorithms} presents the anomaly detection algorithm we implemented. The experimental setup and procedure are discussed in Section \ref{sec:experiment}. The results of the case study conducted is presented in Section \ref{sec:results}. Section \ref{sec:discussion} discusses about model optimization through TCN and LSTM. Section \ref{sec:con} concludes the paper and highlights the future work. 

\vspace*{-0.08in}
\section{Anomaly Detection in IoT Time Series Data: A Review}
\label{sec:relatedWork}




Time series data are integral part of IoT systems. A traditional problem in time series data analysis deals with anomaly detection in least amount of time with minimal amount of data in order to predict abnormal behavior or possible cyber attacks in their earliest stages. The anomaly detection problem in time series data is involved with discovering abnormal and unusual patterns in the data stream that do not follow with the expected or known patterns. These abnormal and unusual patterns may represent outliers, noises caused by low quality sensors, shocks in economics data, or simply deviations caused by unknown actors. The problem of detecting anomalies in time series data in conventional machine learning algorithms is treated as a classification or regression problem. As a result, there is a myriad body of research work on demonstrating machine learning classification techniques such as Support Vector Machines (SVM), K-Nearest Neighborhood (KNN), regression, and many others \cite{IoT1}. Here, we review some of the deep learning approaches to anomaly detection using deep learning. Readers who are interested, may refer to \cite{IoT1, IoT2} for a complete list of machine and deep learning techniques applied to IoTs data analysis.


Malhotra et al.\ \cite{IoT9} present an LSTM-based encoder-decoder for anomaly detection. The encoder reads time series data and learns a vector representation of the data; whereas, the decoder reconstructs the initial time series through the vector built. The encoder-decoder model is trained based on regular and normal time series data. The decoder part of the model then computes the error for the new data and captures the likelihood of data being abnormal. They report F score of above $0.84$ for the dataset they studied. 

Zhou et al.\ \cite{IoT18} utilize spatial-temporal features to build a convolutional neural networks for anomaly detection. The authors believe that a combination of spatial and temporal features best fit for the type of application domain (i.e., video sequences of crowded scenes) they intended to model. 

Kim et al.\ \cite{IoT8} propose Squeezed Convolutional Variational AutoEncoder (SCVAE) model where input data (i.e., time series data captured by sensors) are expressed through two-dimensional image data; whereas, the output is the mean and variance of the Gaussian distribution of the output produced by the encoder. In this model, the encoder is presented by a CNN structure containing four layers. Similarly, the decoder is modeled using CNN structure of five layers. The authors report an accuracy of $0.962$ using their proposed model. 

Su et al.\ \cite{IoT11} introduce a stochastic recurrent neural network multivariate anomaly detection for time series data called OmniAnomaly. The deriving idea is to learn the normal patterns of multivariate time series using stochastic variable connection and then use the reconstruction probabilities to identify anomalies. The key feature of OmniAnomaly is its capability in providing interpretation using the reconstructed probabilities of the univariate time series. The F-score reported by the authors is $0.89$.

Using Long Short-Term-Memory (LSTM) networks, Li et al.\ \cite{IoT4} build a model using Generative Adversarial Networks (GANS) to capture the temporal correlation of time series. In the proposed Multivariate Anomaly Detection with GAN (MAD-GAN) model, the authors used an anomaly score called DR-score to identify anomalies by discrimination and reconstruction. The authors report F score of $0.77$. 

There are also several other work for addressing the anomaly detection problem in which graph-based modeling \cite{IoT19} and self-attention network \cite{IoT13} are presented. In these models, the structure of graphs are learned through which the anomaly are distinguishable from normal time series data. For instance, He and Zhao \cite{IoT21} report the results of a study for anomaly detection in time series data using TCN where prediction errors are fitted by multivariate Gaussian distribution and then used them to compute the anomaly scores. He and Zaho did not compare the performance of TCN in comparison with some other deep learning-based anomaly detection raising the question whether TCN outperforms other techniques.

In this paper, we also adapt TCN for anomaly detection. However, to have a better insights of the performance of these deep learning-based models,  we compare its performance with those reported by Chen et al.\ \cite{IoT19} as well as the performance of LSTM-based models.

\vspace*{-0.08in}
\section{Background}
\label{sec:background}

\subsection{Long Short-Term Memory (LSTM)}
\label{sec:LSTM}

LSTM, a recurrent deep neural network (RNN), predicts a sequence of data using feedback connections and loops in the network and is able to learn long-term dependencies\cite{lstm1997}. 
Generally speaking, LSTM consist of a chain of repeated cells (i.e. main layers or moduals). 
Each cell in an LSTM model typically consists of three interacting layers including ``forget gate layer'', ``input gate layer'' and ``output gate layer''. These gates are made of a sigmoid neural net layer followed by a point-wise multiplication operation, which enables the LSTM model to remove or add data from or to the flow of information through each cell.
In each LSTM cell, the ``forget gate layer'' is the initial layer. This gate determines whether parts of the information are disregarded. The ``input gate layer'' followed by a ``tanh'' layer filters and updates the new input data and produces the update for the cell state. The ``tanh'' layer then aggregates the data from the previous layers and updates each state value. Finally, the last layer called the ``output gate layer'' generates the output based on the cell state.

\subsection{Temporal Convolutional Network (TCN)}
\label{sec:TCN}


Temporal Convolutional Networks (TCNs) are a type of time-series model that overcomes previous constraints by capturing long-term trends via a hierarchy of temporal convolutional filters. TCNs are divided into two categories: 1) Encoder-Decoder TCN (ED-TCN) uses only a hierarchy of temporal convolutions, pooling, and up-sampling while successfully capturing long-range temporal patterns. The ED-TCN has a few levels (three in the encoder), but each layer is made up of a series of long convolutional filters. 2) Instead of pooling and up-sampling, a dilated TCN uses dilated convolutions and adds skip links across layers\cite{IoT22}.


A dilated convolution is a convolution in which a filter is applied across a larger region than its length by skipping input values with a specified step \cite{IoT24}. In time series, the model must memorize enormous amounts of data, which the causal convolution cannot handle, but a dilated convolution with a bigger filter created from the original filter by dilation with zeros is far more efficient to do the task.



\vspace*{-0.08in}
\section{Anomaly Detection Algorithm}
\label{sec:algorithms}

Algorithm \ref{alg:anomaly} lists a general procedure for anomaly detection. In Algorithm \ref{alg:anomaly}, first data is split into train and test data in a way that the train data contains only normal data and the test data includes a combination of anomalies and normal data. Next, the desired deep model designed for time series analysis is built and trained based on training data. 
 Having the test and predicted values, if the distance between the predicted and observe values is beyond a threshold value, the data point is classified as an anomaly. Finally, based on the classification results, the confusion matrix and then classification metrics are computed and reported.

\begin{algorithm}[h!]
    \caption{Anomaly detection using deep learning models.}
    \label{alg:anomaly}
    \begin{algorithmic}[1]
        \Input
            \Desc{$data$}{multivariate time series data}
            \Desc{$f_{a}$}{input with anomaly}
            \Desc{$fa_{class}$}{true label of each data point for input with anomaly}
            \Desc{$arch$}{deep model architecture (e.g., TCN, LSTM)}
            \Desc{$thr$}{classification threshold}
        \EndInput
        \Output
            \Desc{$f_{p}$}{prediction on feature with anomaly}
            \Desc{$fp_{class}$}{predicted class of each data point for feature with anomaly}
            \Desc{$CM$}{Confusion Matrix}
            \vspace{0.15in}
        \EndOutput
        \hrulefill
        \State $data_{train}, data_{valid}, data_{test} = split\_dataset(data)$
        \State $Model \gets build\_model(arch)$
        \State $Model \gets trainValidate\_model(data_{train},data_{valid})$
        \State $f_{p} = predict(Model,data_{test})$
        \For {$i \in range(1, f_{a})$}
            \If{$abs(f_{p}[i] - f_{a}[i])> Threshold$} \\
                ~~~~~~~~~$fp_{class} = Anomaly$
            \Else
                $fp_{class} = Normal$
            \EndIf
        \EndFor
        \State $CM = confusion\_matrix(fp_{class},fa_{class})$
        \State $precision,recall,F1=compute\_metrics(CM)$
        \State \Return $CM,precision,recall,F1$
    \end{algorithmic}
      \vspace{5px}
\end{algorithm}

The Algorithm starts off with splitting the entire dataset into training, validation, and testing subsets. The training and validation datasets do not include any anomaly data and they are utilized for building and fitting a model; whereas, the testing dataset includes anomaly and normal data. Once the datasets are formed, the deep learning-based models are built and then trained with training and refined with the validation data sets, respectively. Once the best model with minimal loss values are obtained, the built model is tested against the testing dataset and produces the ``predicted'' values. The predicted and actual values are then compared and if the difference between the two is greater than $k$ times of the standard deviation of the predicted data then, it is considered as an anomaly. In our study, we empirically found that $k = 1.75$ works the best for the dataset we studied.

\vspace*{-0.08in}
\section{Experimental Procedure}
\label{sec:experiment}

\subsection{Data Set}
We utilized a part of SWaT dataset (July version 2, 2019 dataset with anomaly) from iTrust, Centre for Research in Cyber Security, Singapore University of Technology and Design \cite{itrustdatasets}. The data in this dataset is collected from Secure Water Treatment (SWaT) which is a water treatment testbed for cybersecurity researchers who aim to design secure Cyber Physical Systems. The dataset consists of 3 hours of SWaT running under normal operating condition and 1 hour with attack incidents. The feature set of this dataset contains network traffic and the values obtained from the sensors and actuators labeled according to their normal/abnormal behaviours.


\subsection{Data Preparation}

\begin{figure}[h]
\includegraphics[width=\linewidth]{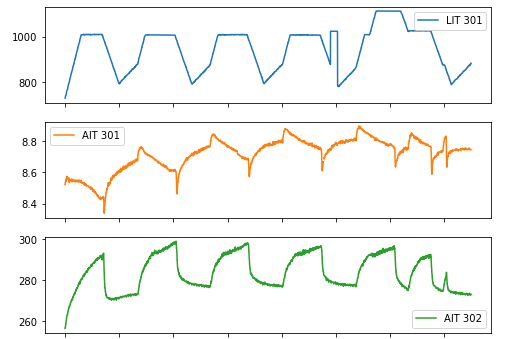}
\caption{Patterns of three sensor: LIT301, AIT301, AIT302.}
\label{fig:features}
\vspace{-0.15in}
\end{figure}

The section of the data we used for our experiment is collected  from 2019-07-20 04:30:00Z to 2019-07-20 08:39:59.004013Z consisting of a total of $14,996$ data points. This partition of data contains the data when LIT301 sensor was under attack for 4 minutes and 32 seconds including records from 9900 (time index: 2019-07-20T07:15:02.0050048Z) to 10170 rows (time index: 2019-07-20T07:19:34.0050048Z). In addition, only 3 sensor data including the data captured by sensors LIT301, AIT301, AIT302 were used for this experiment. Figure~\ref{fig:features} illustrates the patterns of these sensors. These three sensor data are then used as inputs for the multivariate analysis conducted here. 

We scaled the data using standardization in the pre-processing stage.
87.4\% of the data was used for training and validation; whereas 12.6\% of data allocated for testing. 

\subsection{Model Building}

The TCN-based model is implemented using Keras with the Tensorflow backend using 84 filters each with $5 \times 5$ kernel size, with 'same' padding and 'tanh' activation function followed by a regression layer. Furthermore, validation takes place every 50 steps and epoch is set to 100. The batch size is set to 128. The mean squared error is used as the loss function for prediction and 
Adam~\cite{Kingma2015} is adopted as optimization strategy.

The LSTM implementation is based on the Keras Library. We constructed two versions of the LSTM models, one with only one hidden layer and a second one with three hidden layers. The architecture with one layer consists of 64 units. The first, second, and third layers of the three hidden layers architecture contain units of 64, 45, and 35, respectively. The activation function is 'tanh' and the dropout rate is $0.2$ for both architectures to avoid possible overfitting problem.

We built {\it multivariate} models based on TCN and LSTM with LIT 301, AIT 301, and AIT 302 as input data and LIT 301 (the sensor that is under attack) as output. 

\subsection{Assessment Metrics}

 We evaluated the performance of our model using precision, recall and $F_1$ measure, which are common metrics for anomaly detection.
These metrics can be computed through confusion matrix containing counts of samples classified as True Positive (TP), True Negative (TN), False Positive (FP), and False Negative (FN):

\vspace*{-0.15in}
\[
    Precision = \frac{TP}{TP + FP}
\]
\[
   Recall = \frac{TP}{TP + FN}
\]
\[
    F_1 = \frac{2 \times Precision \times Recall}{Precision + Recall}
\]

\vspace*{-0.08in}
\section{Results}
\label{sec:results}

\begin{figure}[h!]
 \vspace{-0.1in}
  \center
  \begin{subfigure}{\linewidth}
      \center
\includegraphics[width=\linewidth, height=4cm]{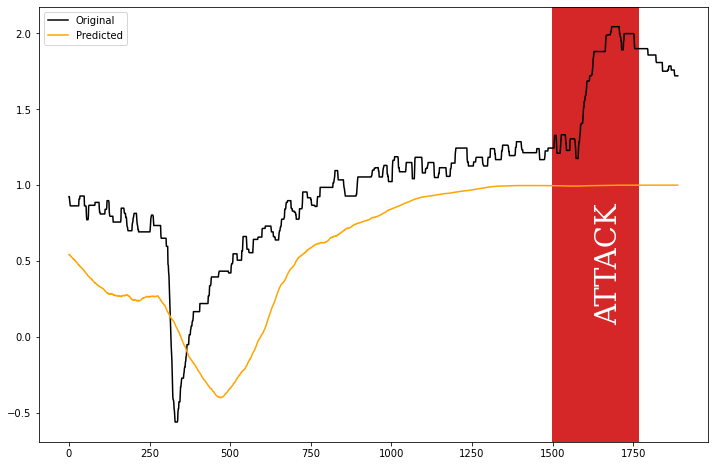}
\caption{LSTM with three Layers.}
\label{fig:lstm-predict}
    \end{subfigure}%
    
           \begin{subfigure}{\linewidth}
      \center
\includegraphics[width=\linewidth, height=4cm]{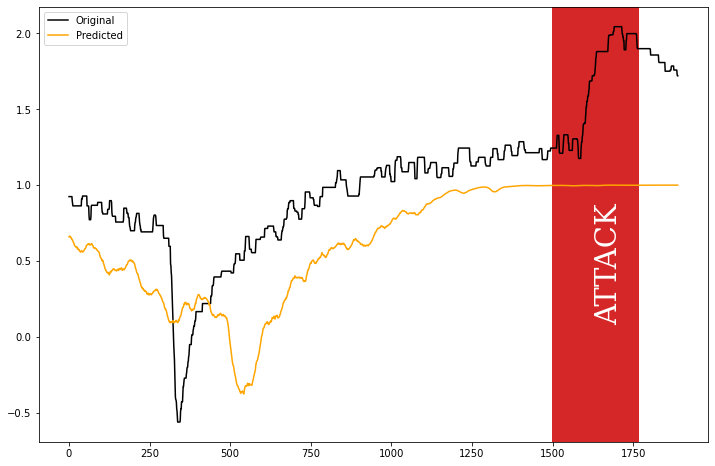}
\caption{TCN with three layers.}
\label{fig:TCN-predict}
    \end{subfigure}%
     \caption{Prediction plots for AIT301.}
  \label{fig:prediction-plots}
  \vspace{-0.15in}
\end{figure}

The experiments are conducted on a MacBook Pro (16-inch, 2019) with 2.3 GHz 8-Core Intel Core i9 processor and AMD Radeon Pro 5500M 4 GB graphics. Once the LSTM and TCN models are built and trained, the prediction plots and predicted evaluation metrics are generated.  Figure~\ref{fig:prediction-plots} illustrates the predicted and actual data using LSTM and TCN model, each with 3 hidden layers, where x-axis holds time; whereas, the y-axis is the standardized data. The plot demonstrates the prediction and actual data for AIT301 sensor. In each plot, 
The predicted values are represented in yellow; whereas, actual values are plotted in black color. The figure shows that the line belonging to the predicted values closely follows the black data until the attack occurs. 

In order to decide whether any anomaly has occurred, we follow the well-known technique in taking into account the distance between the observed data and the mean of predicted values (i.e., $k$ standard deviation rule for outliers). According to our exploratory analysis, we observed that $ k = 1.75$ standard deviation away from the mean of predicted values yielded the best results. Therefore, we set the $1.75 \times SD$ as the threshold to decide about a data point being abnormal/normal.

We classified the actual data to two disjoint classes of abnormal and normal. Given the classified actual and predicted data points to abnormal and normal, the confusion matrix for our experiment produced by the TCN model with 3 hidden layers is as follow:

\vspace*{-0.15in}
\[
TCN_{3hidden}\;Confusion\;Matrix=
\begin{bmatrix}
\label{eq:tcn-cm}
1546   & 166 \\
99   & 78 
\end{bmatrix}
\]

According to the confusion matrix obtained for TCN model, True Positive (TP),  True  Negative  (TN),  False  Positive  (FP),  and  False Negative (FN) are $1546$, $78$, $99$, and $166$, respectively. Using this confusion matrix, the classification performance metrics are computed. Table~\ref{tbl:eval} shows the performance metrics. The TCN model achieves $0.939$, $0.903$, and $0.920$ for Precision,  Recall, and F1, respectively. To demonstrate the effectiveness of the model, the results are compared with the results from Graph Learning with Transformer for Anomaly detection (GTA) model introduced  in~\cite{IoT19}. Table~\ref{tbl:eval} also reports the performance metrics and their values for GTA where Precision, Recall, and F1 scores for GTA model are computed as $0.740$, $0.960$, and $0.840$, respectively. We observe that TCN model outperforms GTA with better performance. It is also important to note that the GTA model is based on the whole SWaT data set; whereas, our study is based on some portion of the SWaT.

Performing a similar experiment, the confusion matrix using the LSTM model with only 1 hidden layer is formed as follow:

\vspace*{-0.15in}
\[
LSTM_{1hidden}\;Confusion\;Matrix=
\begin{bmatrix}
\label{eq:lstm-cm}
1403     & 309 \\
85   & 92 
\end{bmatrix}
\]

According to this matrix, the True Positive (TP),  True  Negative  (TN),  False  Positive  (FP),  and  False Negative (FN) are $1403$, $92$, $85$, and $309$, respectively. Hence, the LSTM model achieves $0.942$, $0.819$, and $0.875$ for Precision,  Recall, and F1, respectively for the LSTM model with 1 hidden layers. We observe that the LSTM model with 1 hidden layer improves the Precision. However, the values for Recall and F1 scores are reduced. It is also important to consider the time needed to train both models. The training time for TCN model is around 5:55:50; whereas, the training time for the LSTM model with 1 hidden layer is around 2:43:29. A trade off analysis would be needed when deciding about the best model to adapt for the anomaly detection problem. 

Changing the architecture of the LSTM model to 3 hidden layers and replicating a similar experiment, the confusion matrix for the LSTM model with 3 hidden layers obtained is as follow:

\vspace*{-0.15in}
\[
LSTM_{3hidden}\;Confusion\;Matrix=
\begin{bmatrix}
\label{eq:lstm-3hl-cm}
1480     & 232 \\
89   & 88 
\end{bmatrix}
\]

According to this matrix, the True Positive (TP),  True  Negative  (TN),  False  Positive  (FP),  and  False Negative (FN) values are $1480$, $88$, $89$, $232$, respectively. Hence, the LSTM  model with 3 hidden layers achieves $0.943$, $0.864$ and $0.901$ for Precision,  Recall and F1, respectively.

\begin{table}[h]
\caption{Evaluation.}
\begin{center}
\begin{tabular}{l|ccccc}
\hline
\multicolumn{1}{c|}{Model} & No. Hidden & Precision & Recall & F1 & Training  \\ 
\multicolumn{1}{c|}{} &  Layers &  &  &  &  Time \\ 
\hline
TCN & 3 & 0.939 &0.903 & {\bf 0.920} & 5:55:50 \\ \hline
LSTM & 1 & 0.942 & 0.819 & 0.875 &2:43:29 \\ \hline
LSTM & 3& 0.943 & 0.864 & 0.901 & 6:28:01 \\ \hline
GTA~\cite{IoT19} & -- & 0.740 & 0.960  & 0.840 & - \\ \hline
\end{tabular}%
\end{center}
\label{tbl:eval}
\end{table}

In Table~\ref{tbl:eval}, we observe that both TCN and LSTM with 3 hidden layers perform comparably with TCN performing slightly better. The LSTM model with 3 hidden layers performs better in Precision; whereas, the TCN model performs better for Recall and F1 score. Similarly, a trade off analysis would be needed since the  LSTM with 3 hidden layers require more time to train (i.e., 6:28:01).

\vspace*{-0.08in}
\section{Discussion}
\label{sec:discussion}

For evaluation and  comparison, we consider F1 as the main metric since Recall shows the proportion of true anomalies which are detected, Precision shows the proportion of detected anomalies that are true and F1 Score shows the overall performance of the anomaly detection by using both Recall and Precision.
Considering F1, TCN model outperforms the other models ($F1 = 0.920$), as it is shown in Table~\ref{tbl:eval}. Increasing the number of LSTM's hidden layers from one to three, the performance of the LSTM model increases from 0.875 to 0.901 for F1 score. On the other hand, the training time increases as well in a way that the training time for the LSTM model with 3 hidden layers is almost three times higher than the training time for the LSTM model with only one hidden layer. The training time for the LSTM model with 3 hidden layers is somewhat higher than the training time for the TCN model with 3 hidden layers. 

\begin{figure}[h!]
  \center
  \begin{subfigure}{0.5\linewidth}
      \center
\includegraphics[width=\linewidth, height=4cm]{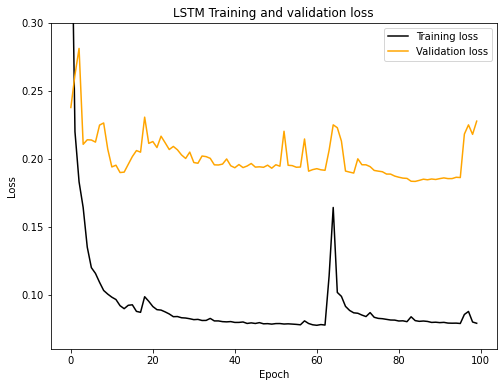}
\caption{LSTM}
\label{fig:lstm-loss}
    \end{subfigure}%
  \begin{subfigure}{0.5\linewidth}
  \center
\includegraphics[width=\linewidth,
height=4cm]{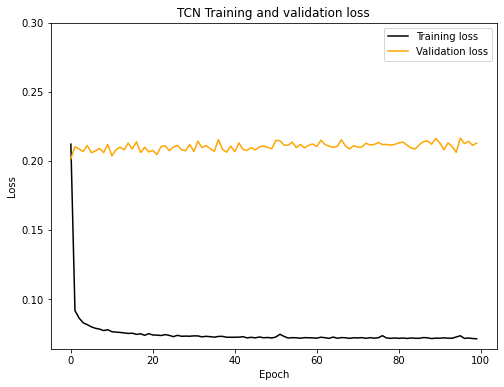}
\caption{TCN}
\label{fig:tcn-loss}
    \end{subfigure}%
     \caption{Loss values for the LSTM and TCN models with 3 hidden layers.}
  \label{fig:loss}
  \vspace{-0.15in}
\end{figure}

The training and validation loss functions for the LSTM model with 3 layers and the TCN model with 3 hidden layers are shown in Figure~\ref{fig:loss} where x-axis shows the number of epoch; whereas, the y-axis holds the loss values. By looking at the elbow joint points in training loss values (i.e., the point where increasing the number of epoch makes only a small difference in the loss function), it is apparent that TCN reaches to a stable model faster with less error in fewer numbers of epoch. This is justifiable since the convolutional layers in TCN improve the learning process in each epoch.

Both convolutional layers in TCN and Recurrent layers in LSTM make the learning process more time consuming that matches our training time for TCN and LSTM shown in Table~\ref{tbl:eval}. As it is reported in Table~\ref{tbl:eval}, for the same number of epochs  (i.e., $Epoch = 100$), TCN model takes longer (i.e., training time) for training compared to the LSTM model with only one hidden layer and almost as the same amount of time as the training time for LSTM with 3 hidden layers. In addition, given the validation loss functions for both models, the validation loss function for TCN is close to optimal in early epochs and also smoother with less oscillation compared to LSTM model with three hidden layers.

Comparing the performance metrics of GTA~\cite{IoT19} with our models, 
it is noticeable that all the models in our experiment overcome the GTA model considering the F1 score since they all achieve  F1 score greater than $0.840$ computed by the GTA model.
The Precision and F1 metrics for GTA~\cite{IoT19} are much lower than the TCN and LSTM models. Recall is the only metric that GTA~\cite{IoT19} performs better than both LSTM models with one hidden layer and 3 hidden layers and a small outdo compared to the TCN model.




\vspace*{-0.08in}
\section{Conclusion and Future Work}
\label{sec:con}

Time series data play an integral role in many application domains including IoTs. Due to the additional features such as temporal properties as well as uncertainty in these types of data, building prediction models with high accuracy is a grand challenge. In particular, when sensors are of low quality and it becomes hard to decide whether an anomaly occurred due to noisy data captured by the sensor or it is the result of tampering caused by cyber attacks. 

Several traditional regression-based methodologies such as Auto-Regressive Integrated Moving Average (ARIMA) and conventional machine learning approaches (e.g., Support Vector Machines (SVM)) have been applied for modeling the aforementioned problem. The recent advancement in deep learning modeling has made these sophisticated approaches available for addressing the prediction and anomaly detection in time series data. 

In our previous research \cite{ICMLA2018, BiLSTM}, We compared the performance of ARIMA and LSTM and showed that LSTM outperformed the state-of-the-art in time series modeling (i.e., ARIMA) substantially. In the present work, we compare the performance of LSTM and its counterpart TCN in terms of precision and training time. Given the architectural differences in these two models the results will be of utmost interest when building prediction models or detecting anomalies. 

LSTM is a Recurrent Neural Network (RNN) capable of capturing temporal properties by remembering the previously observed data. On the other hand, TCN is a Convolutional Neural Network (CNN) that through some stack of layers captures the temporal properties of time series data. 

In this work, we built three models: 1) a TCN model with 3 hidden layers, 2) an LSTM model with only 1 hidden layer, and 3) an LSTM model with 3 hidden layers. The intuition of building these models is to compare the performance of TCN with a simple LSTM (i.e., only 1 hidden layer) and also with a LSTM model whose architecture is close to the TCN model (i.e., 3 layers). 

We trained, validated, and tested the models on SWaT, a water treatment testbed for research and training on ICS security \cite{IoT17}. We treated the problem as a classification problem where the objective is to determine whether an anomaly or attack is observed. We then captured the precision, recall, and F1 as well as the training time for all three models. 

We observed that all three models (even the simple LSTM model with 1 layer) perform competitively. In terms of precision all three models achieve at least $0.94$; whereas, in terms of recall the TCN model, the LSTM model  with 3 hidden layers, and the LSTM model with a simple 1 hidden layer achieve $0.903$, $0.864$, and $0.819$.

In terms of F1 scores, we observed that TCN model reports an F1 score of $0.920$ followed by the LSTM model with 3 layers (i.e., $0.901$). The F1 score reported by the GTA model is by no means even close to the scores reported by TCN and LSTM models. 


An interesting observation is demonstrated in Figures \ref{fig:loss} where we noticed that the TCN-based model builds the prediction model faster and thus there is no need for additional training through increasing epochs; whereas, the LSTM model demonstrated some instability in terms of loss. 

The results we observed in this research work favor slightly TCN-based models in some aspects. First, TCN models report better performance in comparison with LSTM and GTA models. Second, TCN models need fewer iterations for training and obtaining the best model. More specifically, the TCN-based model builds the prediction model faster and it is more stable. Further research studies are needed to replicate the experiments conducted in this paper comparing different properties of these two deep learning-based modeling approaches. It is also interesting to study the performance of a hybrid modeling approach such as ConvLSTM \cite{ConvLSTM}.

\vspace*{-0.08in}
\section*{Acknowledgement}
This research work is supported by National Science Foundation (NSF) under Grant No. 1821560.


\bibliography{references}{}
\bibliographystyle{plain}
\end{document}